\documentclass{article}

\usepackage{arxiv}
\usepackage{amsmath,amssymb,amsfonts}
\usepackage{algorithmic}
\usepackage{graphicx}
\usepackage{textcomp}
\usepackage[utf8]{inputenc}
\usepackage{siunitx}
\usepackage{multirow}
\usepackage{tikz}

\newcommand\copyrighttext{%
  \footnotesize \textcopyright 2020 IEEE. Personal use of this material is permitted.
  Permission from IEEE must be obtained for all other uses, in any current or future 
  media, including reprinting/republishing this material for advertising or promotional 
  purposes, creating new collective works, for resale or redistribution to servers or 
  lists, or reuse of any copyrighted component of this work in other works. \\
This paper was accepted for publication in the 25th International Conference on Pattern Recognition (ICPR2020). 
 }
\newcommand\copyrightnotice{%
\begin{tikzpicture}[remember picture,overlay]
\node[anchor=south,yshift=10pt] at (current page.south) {\fbox{\parbox{\dimexpr\textwidth-\fboxsep-\fboxrule\relax}{\copyrighttext}}};
\end{tikzpicture}%
}
\begin{document}
\copyrightnotice

%
\title{Electroencephalography signal processing based on textural features for monitoring the driver’s state by a Brain-Computer Interface}

\author{Giulia Orrù, Marco Micheletto, Fabio Terranova, Gian Luca Marcialis\\
Department of Electrical and Electronic Engineering\\ University of Cagliari, Italy\\
{\tt\small fabio.terranova@mail.com, \{giulia.orru, marco.micheletto, marcialis\}@unica.it}}

\maketitle

\begin{abstract}

In this study we investigate a textural processing method of electroencephalography (EEG) signal as an indicator to estimate the driver's vigilance in a hypothetical Brain-Computer Interface (BCI) system.

The novelty of the solution proposed relies on employing the one-dimensional Local Binary Pattern (1D-LBP) algorithm for feature extraction from pre-processed EEG data. From the resulting feature vector, the classification is done according to three vigilance classes: awake, tired and drowsy. The claim is that the class transitions can be detected by describing the variations of the micro-patterns' occurrences along the EEG signal. The 1D-LBP is able to describe them by detecting mutual variations of the signal temporarily ``close'' as a short bit-code. 

Our analysis allows to conclude that the 1D-LBP adoption has led to significant performance improvement. Moreover, capturing the class transitions from the EEG signal is effective, although the overall performance is not yet good enough to develop a BCI for assessing the driver's vigilance in real environments.
\end{abstract}

\section{Introduction}
In the broad spectrum of cognitive states, vigilance plays an essential role in most human activities. Many working occupations require high levels of alertness to prevent and avoid potential dangers to the labourer.
An example of such activities is driving: drivers, in any circumstance, are required to be active to maintain a correct vehicle trajectory, avert casualties, and react to random dangerous episodes. Research studies show that in a considerable percentage of road accidents, sleep and fatigue have been stated to be contributing factors \cite{horne1995sleep, sagberg1999road}.
 A system capable of detecting when the driver is getting tired, or even falling asleep, and of giving real-time feedback could be a huge benefit in the prevention of both non-fatal and fatal crashes.
 \\
Electroencephalography (EEG) is a monitoring method based on the principles of electrophysiology used to record brain activity.
In the transition between wakefulness and sleep, the EEG is one of the most indicative bio-signals that reflect the event, for the very reason of being a direct indication of those events \cite{berka2007eeg}. Its core characteristic makes it a promising indicator of vigilance and fatigue levels.

Starting from Ref. \cite{zheng2017multimodal}, which developed a multimodal approach for vigilance estimation regarding temporal dependency and combining EEG and forehead EOG in a simulated driving environment, we wanted to exploit the temporal evolution of the mental state during the transition from an \textit{awake} to a \textit{drowsy} state, to detect the event and alert the driver.

We hypothesize that these transitions can be represented by a sequence of a set of bit-codes computed by the one-dimensional version of the Local Binary Pattern operator (1D-LBP, \cite{ojala2002multiresolution}). This work's novelty lies in the use of the 1D-LBP for feature extraction from the EEG signal in a driver's state monitoring system. The analysis is done over a given time-window, named epoch. The goal is to identify the epochs where the subject is getting tired or falling asleep and send a signal (e.g. an audio alarm) to wake the driver up. 
The second point is to investigate if this representation can generalize the drowsiness detection to any user or requires user-specific settings. 

The white-box nature of the proposed method, unlike deep-learning approaches, allows to provide an explanation of the BCI's behavior. Moreover, to make the system response more explainable, we propose new performance metrics that can help standardizing comparisons and evaluate the temporal response of the classifier. 

We focused on three basic properties: the ability to detect a certain class transition, the percentage of cases where the transition is detected at the time reported by the ground truth, and the average delay time in detecting the transition occurrence. The comparison of the reported performance with one of the most effective methods at the state-of-the-art is also performed. Experiments are carried out on the sole dataset publicly available and specifically collected for drowsiness detection, namely, the SEED-VIG data set \cite{zheng2017multimodal}. This data set is very recent. Consequently, the majority of published works used home-made data, thus it is not easy to make the exact point on current achievements. For this reason, we believe that the SEED-VIG data set should deserve more attention from the scientific community.

The rest of this paper is organized in the following sections.  Section \ref{chap:state-of-the-art} provides an overview of the state of the art. Section\ \ref{proposed} describes the purpose of the experimentation. The experimental methodology is presented in Section \ref{exprotocol}. Experiments are reported in Section \ref{results}, while Section \ref{conclusion} summarizes the main results of this paper and offers concluding remarks.

\section{State of the art}

\label{chap:state-of-the-art}
Many methods for the analysis of vigilance levels have been proposed in the past. Early studies focused on image processing techniques by analyzing driver's physical changes, such as eye-closure degree, eyelid movements, head pose, and gaze direction \cite{ueno1994development,boverie1998intelligent,ji2002real,matsumoto2000algorithm}.
These visual-behaviors are usually recorded by one or more video cameras or thermographic cameras to avoid problems in poor or very bright lighting conditions \cite{bergasa2006real}. Another line in active safety system development employs indirect vehicle characteristics like speed, lateral position, and steering angle as driver’s alertness state. Although these methods can achieve a satisfactory level of accuracy, their performance may vary in different environmental situations and driving conditions, and it also may require a significant amount of time to analyze user behaviors.

In this respect, the most accurate indicators are physiological signals such as EEG, electrooculography (EOG), electrocardiography (ECG), and electromyography (EMG). Among them, EEG-based methods appear to be promising in detecting sleep onset while driving. EEG recordings contain information that varies with drowsiness, arousal, sleep, and attention \cite{santamaria1987eeg}.
Wireless and wearable EEG systems have been tested in virtual driving environments to evaluate and get feedback on driving performance \cite{lin2014wireless, lin2013can}. In fact, through the analysis of EEG signals, it is possible to discover a lapse of disengagement of sustained attention \cite{davidson2007eeg}.
This detection is based on the examination of the temporal dynamics of the signal that depicts specific and recognizable patterns up to 20 seconds before the actual error\cite{o2009uncovering}. 
Many previous studies focus on the best way to discriminate the transition between wakefulness and sleep.
Yeo \textit{et al. }\cite{yeo2009can} demonstrated that Support Vector Machines are the best classifiers for this task. They used samples of EEG signals from both states to train the model. In \cite{mardi2011eeg}, chaotic feature extraction is employed to create a significant level of difference between sleepiness and alertness in each EEG channel.
One more approach consists of analysis in the frequency domain, dividing the transformed signal in different EEG rhythms: the main widely adopted intervals are delta (1-\SI{4}{\hertz}), theta (4-\SI{8}{\hertz}), alpha (8-\SI{14}{\hertz}), beta (14-\SI{31}{\hertz}), and gamma (31-\SI{50}{\hertz}) bands \cite{kang}. Each of these bands has a functional meaning: a dense or particular activity in any of these frequency intervals generally corresponds to peculiar cognitive states, physiological processes, and pathologies. In the last decades, various works \cite{LAL2001173, SUBASI2005701} showed that the energy of the feature waves changes with different degrees of fatigue. In particular, an increase in delta, theta, and alpha activity characterizes the EEG signal during driver fatigue.
Therefore, many studies base their fatigue detection systems on the energy analysis of these waves \cite{JAP20092352,bachao}. 
In particular, Zheng \emph{et al.} \cite{zheng2017multimodal} adopted the differential entropy (DE) for features extraction. The ability of DE to discriminate EEG pattern between low and high frequency energy made it suitable to perform both emotion recognition \cite{duan2013differential}, and vigilance estimation. \cite{shi2013differential}.

Other handcrafted features can be computed starting from DE, as differential
asymmetry (DASM) and rational asymmetry (RASM). Nevertheless, as Zheng \emph{et al.} showed in \cite{zheng2015investigating} making a comparison also with the conventional power spectral density (PSD), the DE features are more accurate for this task. 

Relying on these results, we used differential entropy and  \cite{zheng2017multimodal}'s work as a benchmark in our investigation: firstly, because they created a shareable dataset for vigilance estimation (SEED-VIG); secondly, we both investigated the changes in neural patterns associated with vigilance. To the best of our knowledge, this is the most recent and promising state-of-the-art method proposed for this task, whose results can be comparable by a unified experimental protocol.


\section{A system for drowsiness
detection}
\label{proposed}

\begin{figure}[!ht]
	\centering
	\includegraphics[width=0.5\linewidth]{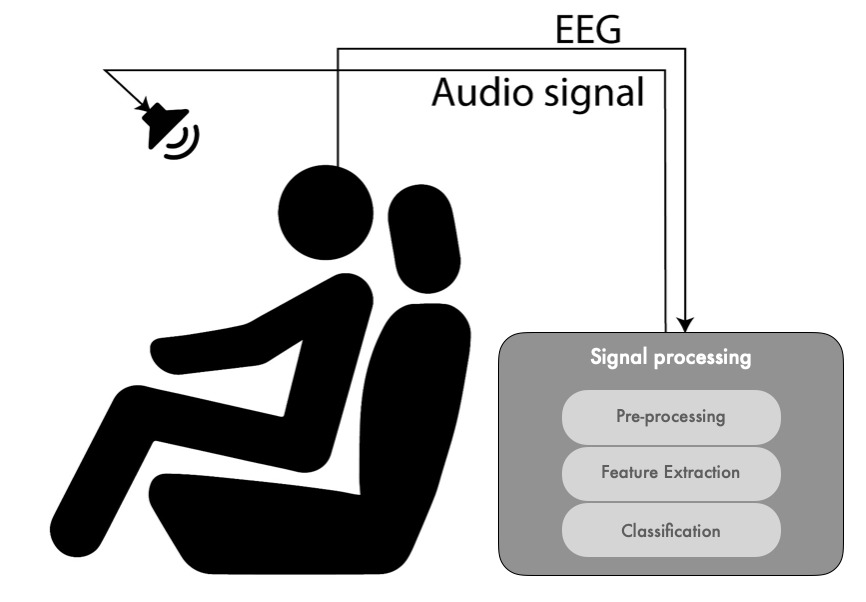}
	\caption{Operating diagram of the BCI system for monitoring the driver's state.}
	\label{fig:bci}
\end{figure}

In this work, we proposed and analysed a hypothetical BCI system to detect drowsiness states and feedback in the driving environment. The BCI is composed of an EEG device capable of recording, a computational unit capable of running the signal processing and
the classification process and a feedback system in real-time to alert the person behind the wheel and wake him up (Fig.\ref{fig:bci}).

Standard measures for BCI efficiency are the accuracy of the classifier installed on the device and temporal analysis of the response of the test classification. In the experiments, we carried out both these criteria.
The classification of the samples was made in three vigilance classes: awake, tired, and drowsy.

The BCI should identify these states with the smallest latency and, secondly, it should generate the least number of false positives (considering the tired/drowsy classes as the positive class), in order to avoid the alarm when the driver is entirely awake.

The novelty of our approach is in the feature extraction algorithm. This is based on the one-dimensional local binary pattern computation for extracting the quantitative histograms from EEG signals. Local Binary Pattern (LBP) \cite{ojala2002multiresolution} is a visual descriptor, primarily used in 2-D image processing for texture segmentation and feature detection \cite{ojala1999unsupervised,he2009quantitative}, whose characteristics can be adapted to work in the one-dimensional (1-D) case. 
The 1D-LBP operator can be defined as a function that takes a signal $x$ as input and examines the neighbour points of $x[i]$ sample, assigning a LBP code to it. The 1D-LBP code is calculated in this form:
\begin{equation}
\resizebox{0.8\textwidth}{!}{
$LBP^{1D}\left(x[i]\right)= \sum ^{\frac{P}{2}-1} _{r=0}\left\lbrace sgn\left(x\left[i+r-\frac{P}{2}\right]-x[i]\right)2^r+ sgn\left(x[i+r+1]-x[i]\right)2^{r+\frac{P}{2}}\right\rbrace$
}
\label{lbp:eq}
\end{equation}
where $sgn(x)$ is the sign function, defined as
\begin{equation}
sgn(x) = \begin{cases}
0 & \text{if } x < 0, \\
1 & \text{if } x \geq 0. \end{cases},
\label{lbp:dec}
\end{equation}
and $P$ is a parameter arbitrarily chosen, that represents the number of neighbouring samples thresholded against the centre one. If the neighbour sample is greater or equal to $x[i]$, its binary code would be 1, if it's lesser, the code is 0. Each binary number is multiplied by a specific binomial weight, relative to the position of the corresponding sample, and summed to obtain the 1D-LBP code that locally describes data. The 1D-LBP operator produces $ 2^P $ different configurations, through whose frequency the analyzed signal is characterized within a given window.

The use of 1D-LBP for signal processing was introduced by \cite{chatlani2010local} in signal segmentation and voice activity detection. To the best of our knowledge, this method has never been applied to the driver's state monitoring.
We have chosen 1D-LBP because, given the properties of the EEG measurements in the drowsiness and sleep states, it could show similarities between windows of signals representing the same transition from a vigilance state to another. As stated by Eqs. \ref{lbp:eq}-\ref{lbp:dec}, this method represents the signal by a sequence of binary strings of $P$ size at each sampling time. This leads to $2^P$ possible configurations of the signal. For a given time window, the histogram of each configuration's frequency represents a specific 'signal state'. We believe that the alteration of the histogram may reflect the alteration of the driver's vigilance from a time window to another. The more the histogram changes, the more abrupt the vigilance state transition.
Therefore, our aim is to investigate whether the configurations where a state transition is present, have different characteristics from those in which there is a state of rest or other types of switching.

In order to verify this claim, we codified each channel of the EEG signal by 1D-LBP over a time window of pre-set size, thus obtaining a sequence of histograms put under classification by a machine learning-based method. We trained two classification models: the linear Support Vector Machine (SVM) and the SVM with a gaussian kernel  (RBF). 
Being the SVM a binary discrimination model and our classification task a multiclass problem, we adopted the one-vs.-one strategy, where each combination of two classes is used for a distinct model training. Hence, we obtained a system made up of $K(K-1)/ 2$ classifiers, where $K$ is the number of vigilance classes in the set $\{ drowsy, tired, awake\}$. $K=3$ in our case, thus:
\[
N_{\text{classifiers}} = \frac{K(K-1)}{2} = \frac{3(2-1)}{2} = 3.
\]
The kernel scale adopted for every Gaussian SVM model was $\sqrt{n}$, being $n$ the size of the feature vector.

Finally, for the sake of comparison with our method, we implemented the method proposed by \cite{zheng2017multimodal}, which consists in extracting the differential entropy (DE) from the EEG signal. Their claim is similar to ours: by the DE extraction, they quantify the amount of uncertainty or randomness of the pattern time: in a state of rest, the signal should be more predictable or repetitive. 

\section{Dataset and pre-processing}

\label{exprotocol}
In order to evaluate the effectiveness of the proposed system, SEED-VIG dataset was used.
SEED-VIG is a freely available multimodal dataset for vigilance estimation published by Zheng et al. in 2017 \cite{zheng2017multimodal}. It consists of EEG and EOG data collected from 23 subjects. All participants had a normal or corrected-to-normal vision and were asked to abstain from assuming caffeine, tobacco, and alcohol before the experiment. Most experiments were performed around 1:30 PM, to induce fatigue easily, and the duration of an entire experiment was approximately 2 hours.

The environment was a simulated driving system: a four-lane highway scene was shown on a LCD screen in front of a real vehicle deprived of unnecessary components (e.g. the engine) and controlled with steering wheel and pedals under the participants' actions. The route was monotonous and primarily straight to induce fatigue more easily in the subjects.

EEG and forehead EOG signals were recorded using Neuroscan system at a \SI{1}{\kilo \hertz} sampling rate (\SI{500}{Hz} bandwidth). The EEG setup recorded 12-channel EEG signals from the posterior site (CP1,
CPZ, CP2, P1, PZ, P2, PO3, POZ, PO4, O1, OZ, and O2)
and 6-channel EEG signals from the temporal site (FT7,
FT8, T7, T8, TP7, and TP8) using the 10-20 system.
Like \cite{zheng2017multimodal}, we split the EEG data into three categories based on the PERCLOS labels, which represents the percentage of eyelid closure over the pupil over time \cite{dinges1998perclos}.
Three vigilance classes, based on the PERCLOS index, were defined:\begin{itemize}
\item \emph{Awake} class: ${PERCLOS} < 0.35$;
\item \emph{Tired} class: $ 0.35 \leq {PERCLOS} < 0.7$
\item \emph{Drowsy} class: $ {PERCLOS} \geq 0.7$
\end{itemize}
A preprocessing was done to filter out any artefact from the EEG data of every experiment: a zero-phase bandpass filtering from \SI{1}{Hz} to \SI{75}{Hz} was applied to the raw data, using a 4th-order Butterworth bandpass filter. The same procedure was done to obtain the frequency bands of interest (\emph{delta}, \emph{theta}, \emph{alpha}, \emph{beta}, \emph{gamma}).
After the pre-processing phase,the 1-D-LPB method \cite{chatlani2010local} and DE method \cite{zheng2017multimodal} were applied for the feature extraction.
The 1D-LBP feature was extracted with parameters $P=2$ and $P=4$ ($P$ is a parameter representing the number of neighbour samples thresholded against the centre one) as a window function of 8 seconds non-overlapping intervals over the EEG signal of every 17 channels for each band. The values of P were chosen small to keep the dimensions of the feature vector contained. Concatenating the feature vector of every channel, we obtained a feature vector for every frame composed of 68 components (4 1D-LBP bins times 17 channels) in the $P=2$ case and 272 (16 1D-LBP bins times 17 channels) in the $P=4$ case.
\subsection{New metrics for temporal response}
To evaluate the classifier's temporal response, we defined new metrics to determine the system's performance.

\textbf{\textit{Hit rate:}}
The \emph{hit rate} is the ratio of successfully recognized class transitions (hits) to the total count of state changes: only one sample is needed to identify a transition and to send an alarm to the driver. We also defined \emph{0-delay hit rate} as the percentage of hits that happen on the first sample of the state change. These values are estimated for two distinct class transitions: \emph{awake to tired} (AT) and \emph{tired to drowsy} (TD). The \emph{awake to drowsy} case was not considered, because it is a very rare event in real applications and this is confirmed by the insignificant number of occurrences in the dataset (the participants go through the intermediate tired state before getting drowsy).
The \emph{hit rate} formula is:
\begin{equation}
    \text{Hit rate}=\frac{H_{AT}}{n_{AT}} + \frac{H_{TD}}{n_{TD}}
\end{equation}
where $H_{\textit{xy}}$ is the \emph{hit rate} for the transition from class \textit{x} to class \textit{y} and $n_{xy}$ the number of transitions.

\textbf{\textit{Mean hit delay:}}
The \emph{hit delay} is defined as the mean of seconds of delay with which the response of the classifier correctly identifies a state change. \emph{Awake to tired} (AT) and \emph{tired to drowsy} (TD) class transitions were considered to obtain this measure.
The \emph{mean hit delay }can be formulated as:
\begin{equation}
\text{Mean hit delay}=\frac{\sum \limits^{n_{AT}} _{i=0} \Delta^{AT}_{i}+\sum\limits ^{n_{TD}} _{i=0}\Delta^{TD}_{i}}{n_{AT} + n_{TD}}
\end{equation}
where $\Delta^{\textit{xy}}= t(hit)_{xy}-t(change)_{xy}$ is the difference between the algorithm detection time of the transition from the class \textit{x} to the class \textit{y} and the actual time of that transition.

\textbf{\textit{False hit rate:}}
The \emph{false hit rate} is the rate of misclassified ``awake'' samples. This value is calculated separately for the \emph{tired instead of awake}, \emph{drowsy instead of awake}, \emph{tired or drowsy instead of awake} cases: 

\begin{equation}
    \text{False hit rate}=\frac{\text{FN}_{(awake)}}{\text{FN}_{(awake)}+\text{TP}_{(awake)}}
\end{equation}
where $FN_{(awake)}$ is the rate of false negative samples and $TP_{(awake)}$ is the rate of true positive samples for the awake class. 
In addition to these new metrics, the accuracy was calculated as:
\begin{equation}
\text{Accuracy}=\frac{\sum \limits^{K} _{i=0} \frac{TP_i+TN_i}{TP_i+FP_i+TN_i+FN_i}}{K}
\end{equation}
where $K$ is the number of classes.
\subsection{Settings optimization}
\renewcommand{\arraystretch}{1.3}
\begin{table*}[!ht]
\centering
\caption{Assessment of the impact of filtering band on system accuracy [\%] using two different SVM classifiers and 1D-LBP features with $P = 2$ and $P = 4$. The accuracies are written as: `mean $\pm$ SD'.
}
{
\begin{tabular}{cc|c|c|c|c|}
\cline{3-6}
 &  & \multicolumn{2}{c|}{\textbf{Linear}} & \multicolumn{2}{c|}{\textbf{RBF}} \\\cline{3-6} 
 & & \textbf{ P=2} &\textbf{ P=4}&\textbf{  P=2} &\textbf{ P=4 }\\ \hline
\multicolumn{1}{|c|}{\multirow{6}{*}{\rotatebox[origin=c]{90}{\textbf{ EEG band}}}} &\textbf{  Delta} & $48.75\pm 0.09$ & $48.31\pm $0.10 & $51.74\pm 0.15$ & $49.30\pm 0.20$ \\ \cline{2-6} 
\multicolumn{1}{|c|}{} & \textbf{ Theta} & $43.73\pm 0.08$ & $44.08\pm 0.09$ & $46.98\pm 0.21$ & $45.21\pm 0.29$ \\ \cline{2-6} 
\multicolumn{1}{|c|}{} &\textbf{  Alpha} & $47.12\pm 0.06$ & $46.58\pm 0.08$ & $52.57\pm 0.10$ & $50.26\pm 0.06$ \\ \cline{2-6} 
\multicolumn{1}{|c|}{} & \textbf{ Beta} & $54.65\pm 0.12$ & $54.42\pm 0.19$ & $60.94\pm 0.17$ & $58.11\pm 0.16$ \\ \cline{2-6} 
\multicolumn{1}{|c|}{} &\textbf{  Gamma }& $51.93\pm 0.08$ & $52.15\pm 0.11$ & $57.80\pm 0.10$ & $55.81\pm 0.18$ \\ \cline{2-6} 
\multicolumn{1}{|c|}{} & \textbf{ Total} & $63.81\pm 0.08$ & $65.02\pm 0.10$ & \boldmath $73.78\pm 0.05$ & $72.95\pm 0.09$ \\ \hline
\end{tabular}%
}
\label{tab:r1}
\end{table*}
We conduct a 5-fold-cross validation process for every EEG band,  to gain a general perspective on how each band could perform in a test process.

Table \ref{tab:r1} shows the results of the different bands analyzed using two different classifiers, a linear SVM model and a SVM model with a Gaussian (RBF) kernel. Features were extracted using 1D-LBP with $P = 2$ and $P = 4$.
The reported results show that the system performs better using the whole band (1-50 Hz) than using single bands. The SVM model with the Gaussian kernel performed better in every case. Moreover, accuracy is higher for the 1D-LBP feature with $P=2$.

This first evidence allows us to point out that:
\begin{itemize}
    \item The vigilance state is not provided by a specific sub-band of the EEG signal. It is a holistic process that requires the contributions of all bands.
    \item The contribution of all bands cannot be described with a linear function, as it can be seen by comparing the accuracy of linear SVM and RBF SVM. As a matter of fact, if the performance difference between these classifiers is not significant when inspecting each EEG sub-band, it is noteworthy when we consider the broadband signal.
\end{itemize}

Therefore, the broadband signal is kept for the rest of the experiments. In order to assess the contribution of each channel, a 5-fold cross-validation was carried out, using $P=2$ for the 1D-LBP extraction and the SVM with RBF kernel ($k = 2$). Results are reported in Table \ref{tab:valchn}.

\begin{table}[!ht]
\centering
\caption{Channels' contribution in terms of mean accuracy [\%] using the 1D-LBP feature extraction ($P=2$) and a SVM classifier with RBF kernel ($k = 2$).}{
\begin{tabular}{lc|c|c|}
\cline{3-3}
 &  & \textbf{5-fold cross-validation accuracy}\\ \hline
\multicolumn{1}{|l|}{\multirow{17}{*}{\rotatebox[origin=c]{90}{\textbf{EEG channel}}}} & FT7 & \boldmath $52.14\pm 0.07$ \\ \cline{2-3} 
\multicolumn{1}{|l|}{} & FT8 & $49.75\pm 0.11$ \\ \cline{2-3} 
\multicolumn{1}{|l|}{} & T7 & $51.54\pm 0.08$ \\ \cline{2-3} 
\multicolumn{1}{|l|}{} & T8 & $50.25\pm 0.06$ \\ \cline{2-3} 
\multicolumn{1}{|l|}{} & TP7 & $50.07\pm 0.08$ \\ \cline{2-3} 
\multicolumn{1}{|l|}{} & TP8 & $48.87\pm 0.13$ \\ \cline{2-3} 
\multicolumn{1}{|l|}{} & CP1 & $47.10\pm 0.07$ \\ \cline{2-3} 
\multicolumn{1}{|l|}{} & CP2 & $46.31\pm 0.14$ \\ \cline{2-3} 
\multicolumn{1}{|l|}{} & P1 & $49.22\pm 0.02$ \\ \cline{2-3} 
\multicolumn{1}{|l|}{} & PZ & $48.87\pm 0.11$ \\ \cline{2-3} 
\multicolumn{1}{|l|}{} & P2 & $47.24\pm 0.10$ \\ \cline{2-3} 
\multicolumn{1}{|l|}{} & PO3 & $49.44\pm 0.06$ \\ \cline{2-3} 
\multicolumn{1}{|l|}{} & POZ & $50.18\pm 0.06$ \\ \cline{2-3} 
\multicolumn{1}{|l|}{} & PO4 & $50.28\pm 0.04$ \\ \cline{2-3} 
\multicolumn{1}{|l|}{} & O1 & $50.87\pm 0.17$ \\ \cline{2-3} 
\multicolumn{1}{|l|}{} & OZ & $51.79\pm 0.14$ \\ \cline{2-3} 
\multicolumn{1}{|l|}{} & O2 & $51.28\pm 0.03$ \\ \hline
\end{tabular}
}
\label{tab:valchn}

\end{table}

We observe that a channel alone cannot achieve the performance of all 17 channels. However, the performance increase from $52\%$ of the $FT7$ channel to $74\%$ of all channels suggests a weak correlation among 1D-LBP features per channel. 
On the basis of these observations, we keep the whole channels for the final BCI design and implementation.

\section{BCI design and test}

\subsection{Experimental protocol}
To test the BCI performance, we carried out two types of tests:
\begin{itemize}
    \item In a first attempt, we train and test the system on the same group of users, a common approach when designing BCI interfaces. We refer to this as ``user-specific'' test.
    \item Secondly, we train the system on a user population that is totally different from that involved for the final use, in order to investigate the system's generalization capability.  We indicate this test as ``generic-users'' test.
\end{itemize}

The dataset was randomly split into two parts, a first part containing $\sim80\%$ of users (18 users) and a second part containing the remaining $\sim20\%$ of users (5 users).
The first part was divided to make a 5-fold-cross validation, and for each iteration after training, the accuracy was calculated on both the validation set (user-specific) and the unknown users from the second part (generic-user).
The experiments were done both with features extracted using the proposed 1D-LBP, and DE features \cite{zheng2017multimodal}.
Each experiment was repeated 20 times and the results are averaged over those runs.

\subsection{Results}
\label{results}
In this section, we reported the results of the experiments carried out to evaluate the applicability of the 1D-LBP method for the feature extraction in a system for drivers' vigilance estimation through the analysis of the EEG signal.
At first, the experiments to evaluate the feasibility of the generic-users vigilance estimation system were made.
We then evaluated how the system responds in the temporal domain, using the metrics introduced in section \ref{exprotocol}. Finally, we evaluate the statistical significance of the experiments performed.

\subsubsection{User-specific and generic-users systems and comparison with the state of the art}
Tab. \ref{tab:r2} shows the generic-user and user-specific system's accuracy. The comparison with the state of the art shows that the best general performance is obtained by the SVM model with RBF kernel trained with 1D-LBP features. The improvement we obtained is relevant by considering the amount of data and the compactness of the 1D-LBP features set.

\begin{table*}[!ht]
\centering
\caption{User-specific and generic user test accuracies [\%]. 1D-LBP is compared to DE in terms of accuracy in classification. Results for generic user tests show how a generic user system could have very low performance.}
{
\begin{tabular}{c|c|c|c|}
\cline{2-4}
 & 1D-LBP (linear) & 1D-LBP (RBF) & DE (RBF) \\ \hline
\multicolumn{1}{|c|}{User-specific} & $65.78 \pm 1.68$ & \boldmath $77.89 \pm 1.17$ & $70.36 \pm 1.32$ \\ \hline
\multicolumn{1}{|c|}{Generic user} & $43.27 \pm 8.46$ & \boldmath $44.50 \pm 8.55$ & $42.36 \pm 9.24$ \\ \hline
\end{tabular}%
}
\label{tab:r2}
\end{table*}

Concerning the generic-users results, both 1D-LBP and DE features achieve a low, unacceptable performance. Actually, the EEG related to fatigue and drowsiness appears stable within an individual across different sessions, but it varies between users because of the individuality degree embedded in any EEG signal \cite{lin2005eeg}.
Indeed, the neurophysiological signals are characterized by strong inter-subjects variations. If we suppose that both 1D-LBP and DE are able to extract partly such individuality, the configurations of 1D-LBP and DE is consequently different even when observing the awake class. This led to a bad parametrization of the decision surface of the adopted classifier. This also explain why EEG is also proposed for personal identification, and in particular the differential entropy as measurement of individuality by \cite{Phung2014UsingSE}. The confirmation of this is given by the correspondent accuracies achieved by the user-specific system.

\begin{figure}[!ht]
	\centering
	\includegraphics[width=16cm]{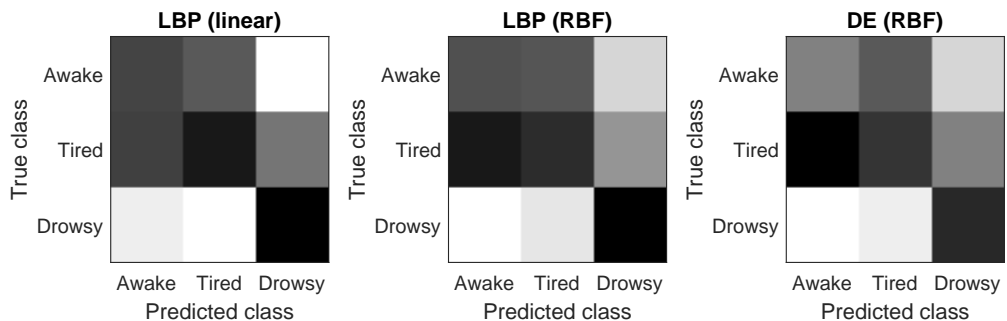}
	\caption{Confusion matrices related to the comparison with the state of the art (DE features) for ``generic-users'' experiments. Each matrix was calculated as the `mean matrix' of all sessions' matrices and is colour coded (greater values, darker colours). For all three analyzed classifiers, the system is perfectly able to distinguish the ``drowsy'' class while the ``tired'' and ``awake'' classes are confused.}
	\label{fig:confmats2}
\end{figure}

However, another point of view must be observed, before going in depth of the user-specific system performance. Fig. \ref{fig:confmats2} shows the confusion matrices of the three SVM models for the generic-users system. We are not interested in `tired' samples being confused with the drowsy class, because the corresponding feedback's alarm of the hypothetical BCI could be used as a precautionary measure in real-world applications. What we are more concerned about is the misclassification of `awake' samples. From Fig. \ref{fig:confmats2}, it is possible to notice that all the classifiers have a high accuracy in the classification of `drowsy' samples while generating confusion in the 'awake' and 'tired' ones. We read this result as an indirect confirmation that each user has its own way of representing, by EEG, a brain state in which he/she is still ''conscious'', that is, he/she has full or partial control of his/her actions. This makes very difficult to classify this cases. Where getting drowsy, this ability is partially lost, and the system becomes more accurate.

\subsubsection{Response dynamics} \label{dyncs}
In this section, we refer to the user-specific system, which show the best accuracy. Therefore, this design approach is the most promising candidate for a real BCI for the driver's vigilance detection.

We go in depth of its performance analysis, because it is not only important how accurate the system is, but also how the system responds in the temporal domain. In fact, what the hypothetical BCI should do is to identify the abrupt changes in the driver's cognitive state with the minimum delay.
Therefore, we defined new metrics to determine the performance of the system in the time domain. These metrics are the hit rate, the 0-delay hit rate, the mean hit delay and the false hit rate (presented in the Section \ref{exprotocol}). Tables \ref{tab:rates1}, \ref{tab:rates2} and \ref{tab:falserates} report the evaluation of these metrics.

First of all, we observe that the LBP-based classifier detects the $96.23\%$ of \emph{awake to tired} transitions with a mean delay of $6.52$ seconds and  $86.75\%$ of \emph{tired to drowsy} transitions with a mean delay of $70.96$ seconds. The \emph{awake to tired} state change is detected with zero delay in the $7.36\%$ of hits. These data suggest that the extracted feature could be used in the case that BCI system has to alarm the driver when he or she is getting tired. However, the false rate relative to the tired class discourages this kind of detection in real-world scenarios, because the $52\%$ of alarms are given when the driver is in the awake state. Consequently, 1 over 2 times the driver would be bored by a useless advice of stop driving because of getting tired. With such a  behaviour of the vigilance system, the risk is that the driver would be lead to turn off it.

Beside this, the reported tables motivates the following observations:
\begin{itemize}
\item the \emph{awake to tired} transition is detected better than the \emph{tired to drowsy} one. This difference could be caused by a higher confusion between the drowsy and tired classes. Due to the natural smoothness of such classes, this could be avoided by the assessment of different PERCLOS thresholds for labeling;
\item the 1D-LBP features perform better than the DE feature. This means that the micro-textural information extracted through the 1D-LBP is more discriminating than the signal entropy, but their performance leads to an indirect confirmation of the respective claims;
\item the \emph{tired to drowsy} state change is recognized with a substantial delay, over a minute, that is unacceptable in real-world applications. In a real application the BCI should therefore rely its operation on the \emph{awake to tired} transition, the only one approaching a reasonable speed to alert the driver;
\item `awake' samples are wrongly marked as `tired' the 52\% of the time. This means that the BCI could alarm the awake driver in a significant number of occurrences. Anyway, this kind of misclassification is much less crucial than the \textit{awake-to-drowsy} or the \textit{tired-to-drowsy} transitions, therefore the system could be based on the latter to generate the alarm and avoid false alarms.
Unfortunately, as mentioned before, the \textit{awake-to-drowsy} transition is infrequent and the \textit{awake-to-drowsy} transition has a mean hit delay above the minute. For this reason, it is possible to state that these limitations make this kind of systems immature for a real application.
\end{itemize}

\begin{table*}[!ht]
\centering
\caption{Hit rate [\%], 0-delay hit rate [\%] and mean hit delay [s] for the \emph{Awake to tired} transition. The detection of this transition is very accurate, especially for the 1D-LPB method.}{
\begin{tabular}{c|c|c|c|}
\cline{2-4}
\emph{Awake $\rightarrow$ tired} & 1D-LBP (linear) & 1D-LBP (RBF) & DE (RBF) \\ \hline
\multicolumn{1}{|c|}{Hit rate [\%]} & $96.20\pm 6.30$ & \boldmath $98.10\pm 2.70$ & $92.40\pm 8.70$ \\ \hline
\multicolumn{1}{|c|}{0-delay hit rate [\%]} & $72.80 \pm 15.10$ & \boldmath $73.40\pm 13.20$ & $70.70\pm 16.80$ \\ \hline
\multicolumn{1}{|c|}{Mean hit delay [s]} & $16.55 \pm 19.97$ & \boldmath $6.52 \pm 5.66$ & $10.35 \pm 8.72$ \\ \hline
\end{tabular}%
}
\label{tab:rates1}
\end{table*}

\begin{table*}[!ht]
\centering
\caption{Hit rate [\%], 0-delay hit rate [\%] and mean hit delay [s] for the \emph{Tired to drowsy} transition. The comparison with the previous table shows that this status change is detected less than the \emph{awake to tired} transition. However the use of the 1D-LBP features with respect to the state of the art considerably increases the detection of this transition.}
{
\begin{tabular}{c|c|c|c|}
\cline{2-4}
\emph{Tired $\rightarrow$ drowsy} & 1D-LBP (linear) & 1D-LBP (RBF) & DE (RBF) \\ \hline
\multicolumn{1}{|c|}{Hit rate [\%]} & \boldmath $86.80\pm 13.40$ & $80.10\pm 18.80$ & $48.80\pm 24.90$ \\ \hline
\multicolumn{1}{|c|}{0-delay hit rate [\%]} & \boldmath $31.10\pm 17.60$ & $20.00\pm 14.10$ & $13.80\pm 16.08$ \\ \hline
\multicolumn{1}{|c|}{Mean hit delay [s]} & $70.96\pm 58.65$ & \boldmath $66.40\pm 43.47$ & $118.64 \pm 98.21$ \\ \hline
\end{tabular}%
}
\label{tab:rates2}
\end{table*}

\begin{table*}[!ht]
\centering
\caption{False hit rates [\%]: the values were calculated for three misclassification cases. The evaluation of this metric confirms what is shown by the confusion matrices. The awake class is more confused with the tired class. The low confusion between awake and drowsy classes is a positive result for the BCI operation.}
{
\begin{tabular}{c|c|c|c|}
\cline{2-4}
\emph{Awake misclassified as} & 1D-LBP (linear) & 1D-LBP (RBF) & DE (RBF) \\ \hline
\multicolumn{1}{|c|}{Tired} & \boldmath $52.25\pm 15.66$ & $56.21\pm 14.10$ & $74.12\pm 15.96$ \\ \hline
\multicolumn{1}{|c|}{Drowsy} & \boldmath $6.52\pm 10.59$ & $8.79\pm 11.55$ & $5.45\pm 7.03$ \\ \hline
\multicolumn{1}{|c|}{Tired or drowsy} & \boldmath $58.77\pm 15.76$ & $65.00\pm 15.63$ & $79.57 \pm 17.14$ \\ \hline
\end{tabular}%
}
\label{tab:falserates}
\end{table*}

\subsubsection{Statistical significance analysis}

In this section, we analyze if the hit rate values calculated for the \emph{awake to tired} and the \emph{tired to drowsy} transitions and the value of false hit rate calculated for the awake class have a statistical significance concerning the sample size. By obtaining a positive answer, our results could be taken into account by the scientific community for further investigations or to confirm or deny our claims by adopting the same data, experimental protocols and evaluation metrics.

In Table \ref{table:samples}, the number of samples of the awake class with respect to the total number of samples and the number of transitions between the awake and tired classes and between the tired and drowsy classes have been reported.
Their statistical significance is assessed by evaluating the \textit{margin of error}, that is, the interval estimation expressing the range of values above and below a confidence interval \cite{stat}.
It is possible to calculate the margin of error by the following:
\begin{equation}
\centering
margin \thinspace of \thinspace error = critical\thinspace value * standard\thinspace error
\end{equation}

When the sampling distribution of a statistic is normal or nearly gaussian, the critical value can be expressed as a t-score or as a z-score. The critical value as a z-score at 95\% level of confidence is $\pm1.96$. 
The standard error ($SE$) can be calculated as:
\begin{equation}
\centering
SE=\sqrt{\dfrac{p(1-p)}{n}}
\end{equation}
in which $n$ is the sample size and $p$ is the population proportion.
The goal was to calculate the margin of error on the hit rate calculated for the \emph{awake to tired} and the \emph{tired to drowsy} transitions and on the false hit rate calculated for the awake class.
  
By considering the samples distribution shown in Table \ref{table:samples}, we have:
$p=7405/20355=0.36$, being the sample size equal to 20355 (23 experiments * 885 samples).
From Eq. (8), we get a standard error equal to $0.0033$ for the awake class.
\begin{table}
\centering
\caption{SEED-VIG dataset composition based on the awake classes and the \emph{Awake $\rightarrow$ tired} and \emph{Tired $\rightarrow$ drowsy} transitions.}{
\begin{tabular}{|l|l|}
\hline
\textbf{Total samples}  & 20355 \\ \hline
\textbf{Awake samples}  & 7405  \\ \hline
\textbf{Total transitions} & 333  \\ \hline
\textbf{\emph{Awake $\rightarrow$ tired}} & 114 \\ \hline
\textbf{\emph{Tired $\rightarrow$ drowsy}} & 60 \\ \hline
\end{tabular}
}
\label{table:samples}
\end{table}

\begin{table}
\centering
\caption{Margins of error referred to the false hit rate calculated for the awake class and to the hit rate calculated for the \emph{awake to tired} and the \emph{tired to drowsy} transitions.}
{
\begin{tabular}{|l|c|}
\hline
\textbf{}                           & \textbf{Margin of error} \\ \hline
\textbf{Awake samples}              & 0.66\%                   \\ \hline
\textbf{\emph{Awake $\rightarrow$ tired}}  & 5.10\%                   \\ \hline
\textbf{\emph{Tired $\rightarrow$ drowsy}} & 4.13\%                   \\ \hline
\end{tabular}
}
\label{table:margin}
\end{table}

The margin of error is 0.66\%, which corresponds to the maximum estimation error under the assumption of gaussian distribution of measurements. Performing the same steps but considering $n=333$, for the \emph{awake to tired} and the \emph{tired to drowsy} transitions, the margin of error is, respectively, 5.10\% and 4.13\% (Table \ref{table:margin}). 
In conclusion, the hit rate results for the \emph{awake to tired} and the \emph{tired to drowsy} transitions are accurate at the 95\% confidence level respectively plus or minus 5 and 4 percentage points. The false hit rate (Table \ref{tab:falserates}) is accurate plus or minus one percentage point. These reported values allow us to consider reliable the performance parameters values obtained in our experiments. 

\section{Conclusions}
\label{conclusion}
In this work, we proposed the 1D-LBP algorithm for assessing the driver's vigilance by the acquisition of the electroencephalography signal (EEG). We dealt with a three-classes classification problem where the assigned classes are awake, tired and drowsy. The novelty of the approach relies on a textural descriptor with the aim of characterizing the transitions from a class to another one by a set of bit-codes over a time window. It was expected that, whilst a transition awake-to-drowsy appeared, the electrodes might point out a decrease in the neural activity devoted to the scene observation, and this variation impact on the frequencies of the bit-codes from a time window to the next one. The one-dimensional LBP (1D-LBP) is a instrument designed to investigate if this is true.

For confirming our claim, we assessed well-known performance parameters and introduced novel ones for the transitions detection and the related time delay. These new temporal metrics allowed us to evaluate accurately the limitations of the proposed method. We compared our results with those of a reference method, based on a similar hypothesis, by adopting the same publicly available data set and the same experimental protocol. Instead, the majority of published works on this topic used home-made data, thus making impossible to assess the level of current achievements.

The proposed method appeared to exhibit strong effectiveness in detecting the awake-to-tired transitions (only 6 sec of delay and the best hit rate, namely $98\%$). Therefore, the claim of differentiating this relevant changes in the vigilance state by 
extracting a set of bit-codes frequencies appeared to be confirmed. 

On the other hand, the proposed method did not exhibit an user-independent performance and still needs user-specific settings to be effective.

Future works will be devoted to improve the performance, especially by reducing the false detection rate, in order to make the implementation of a BCI based on 1D-LBP feasible in real applications.  

\bibliographystyle{abbrv}
\bibliography{ref}
%

\end{document}